\documentclass[sigconf]{acmart}

\usepackage{multirow}

\AtBeginDocument{%
  \providecommand\BibTeX{{%
    \normalfont B\kern-0.5em{\scshape i\kern-0.25em b}\kern-0.8em\TeX}}}

\setcopyright{acmcopyright}
\copyrightyear{2020}
\acmYear{2020}
\acmDOI{****}

\acmConference[]{ACM Conference Template}{Date}{}
\acmBooktitle{ACM Conference}
\acmPrice{**.00}
\acmISBN{***}

\usepackage{adjustbox}

\begin{document}

\title{Knowledge-Aided Open-Domain Question Answering}


\author{Mantong Zhou}
\affiliation{Tsinghua University, China}
\email{zmt15@mails.tsinghua.edu.cn}

\author{Zhouxing Shi}
\affiliation{Tsinghua University, China}
\email{zhouxingshichn@gmail.com}

\author{Minlie Huang}
\authornote{Corresponding author}
\affiliation{Tsinghua University, China}
\email{aihuang@tsinghua.edu.cn}

\author{Xiaoyan Zhu}
\affiliation{Tsinghua University, China}
\email{zxy-dcs@tsinghua.edu.cn}


\begin{abstract}
  Open-domain question answering (QA) aims to find the answer to a question from a large collection of documents.
  Though many models for single-document machine comprehension have achieved strong performance, there is still much room for improving open-domain QA systems since document retrieval and answer reranking are still unsatisfactory.
  Golden documents that contain the correct answers may not be correctly scored by the retrieval component, and the correct answers that have been extracted may be wrongly ranked after other candidate answers by the reranking component.
  One of the reasons is derived from the \emph{independent principle} in which each candidate document (or answer) is scored independently without considering its relationship to other documents (or answers)~\cite{PRP}.
  
  In this work, we propose a knowledge-aided open-domain QA (KAQA) method which targets at improving relevant document retrieval and candidate answer reranking by considering the relationship between a question and the documents (termed as \emph{question-document graph}), and the relationship between candidate documents (termed as \emph{document-document graph}). The graphs are built using knowledge triples from external knowledge resources. During document retrieval, a candidate document is scored by considering its relationship to the question and other documents. During answer reranking, a candidate answer is reranked using not only its own context but also the clues from other documents.
  The experimental results show that our proposed method improves document retrieval and answer reranking, and thereby enhances the overall performance of open-domain question answering.
\end{abstract}



\maketitle

\section{Introduction}
Open-domain Question Answering (QA) aims to find answers in a large collection of documents~\cite{drqa}, such as Wikipedia. Such a setting can be normally decomposed into three subtasks: the first is to retrieve relevant documents, the second is to extract answer candidates from the retrieved documents, and the third is to rerank the answer candidates to identify the correct answer.
With the development of information retrieval (IR) methods and reading comprehension (RC) models, most open-domain QA systems adopt the Retriever-Reader-Reranker pipeline~\cite{r3,reranker}. A retriever scores and ranks relevant documents for a given question, a reader extracts candidate answers in top ranked documents, and a reranker selects answer candidates and determines the final answer.

\begin{figure}[h]
  \centering
  \includegraphics[width=\linewidth]{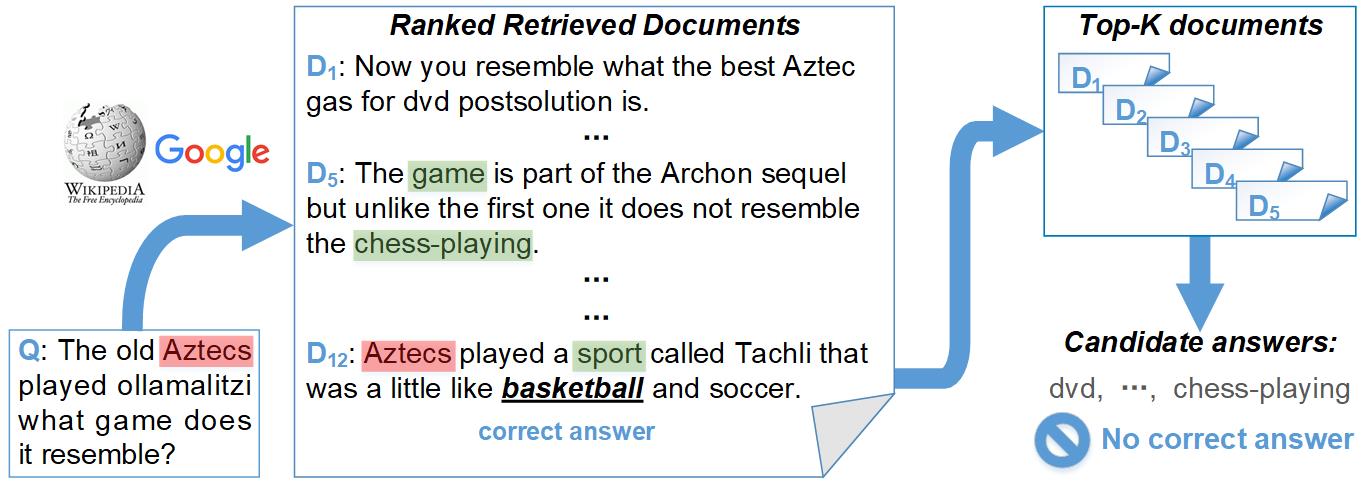}
  \caption{An example from QuasarT~\cite{dhingra2017quasar} where an open-domain QA model, like DrQA~\cite{drqa}, fails to answer the question due to the limitation of the retriever. The golden document ($D_{12}$ in this case) where the correct answer locates is ranked too low (absent in the top 5).}
  \Description{Useful connections between the low-ranked golden document and other corpus.}
  \label{fig:example}
\end{figure}

Though machine reading models (hereafter readers) have achieved strong performance in single document reading comprehension \citep{rajpurkar2016squad:,bert}, these models may obtain sub-optimal performance when dealing with multiple candidate documents because a well-designed document retriever and an answer reranker are critical for the final performance. First,
the documents that contain correct answers (so-called golden documents) may not be correctly retrieved, as can be clearly seen from Figure \ref{fig:example}. In our experiments (see Table \ref{tab:freq} in Section~\ref{sec:pre_exp}), only 59.8\% golden documents can be retrieved in top 5 positions on SQuAD-open~\cite{drqa} by TF-IDF similarity. The situation is even worse on Quasar-T~\cite{dhingra2017quasar} where there are only 48.0\% of golden documents ranked in top 5 positions. Second, correct answers may be discarded by the reranker even though the answers have been extracted by the reader. 
As shown in our experiments on SQuAD-open (Table~\ref{tab:upperbound} in Section~\ref{sec:pre_exp}), there is an increase of more than $15\%$ F1 score in question answering when all answer candidates are input into the reranker.

The major reason can be attributed to the \emph{independent principle} that is widely used in existing work, where each document or answer is modeled independently, without considering its relationship to other candidate documents or answers. 
For document retrieval,
existing open-domain QA systems usually adopt a heuristic retriever based on TF-IDF~\cite{drqa}, BM25~\cite{reranker}, or a neural retriever~\cite{r3,re3}, to score each document. However, existing work only inputs the question and a single candidate document.
The retriever scores each candidate document independently but neglects the rest of candidate documents, thereby producing biased scores~\cite{PRP}. 
As for answer reranking, 
existing work uses neural networks to rerank each extracted candidate merely based on the question and the context around the candidate answer~\cite{re3}.
These reranking models solely aggregate evidence from each answer's context within one document but ignore the clues from other documents.

In this work, we propose to introduce relational knowledge to improve open-domain QA systems by considering the relationship between questions and documents (termed as question-document graph) as well as the relationship between documents (termed as document-document graph). More specifically, we first extract relational graphs between an input question and candidate documents with the help of external knowledge bases, using triples like ``\emph{(chess, type\_of, sport)}'' in WordNet~\cite{wordnet}. 
Then, the document retriever uses question-document and document-document graphs to better retrieve the documents that contain the final answer. 
The answer reranker also leverages such knowledge to evaluate the confidence score of a candidate answer.
By considering the question-document graph, the direct evidence in a document can be used in document retrieval and answer reranking.
Moreover, the document-document graph introduces the global information (from all the other candidates) in addition to the local information (from the current candidate), which helps the retriever/reranker to score candidate documents/answers more accurately.

The contributions of this work are in two folds:
\begin{itemize} 
\item We propose a knowledge-aided open-domain QA (KAQA) model by incorporating external knowledge into relevant document retrieval and candidate answer reranking. We use external knowledge resources to build question-document graph and document-document graph and then leverage such relational knowledge to facilitate open-domain question answering.

\item We evaluate the effectiveness of our approach on three open-domain QA benchmarks (SQuAD-open, Quasar-T, and TriviaQA). 
Experimental results show that our model can alleviate the limitation of existing document retrieval and answering reranking, as well as improve the accuracy of open-domain question answering.
\end{itemize}

\section{Related Works}

\subsection{Open-Domain QA Benchmarks}

Many benchmark datasets have been created to evaluate the ability of answering open-domain questions without specifying the document containing the golden answer.
Quasar~\cite{dhingra2017quasar} requires models to answer a question from top-100 retrieved sentence-level passages. 
SearchQA~\cite{dunn2017searchqa} aims to evaluate the ability of finding answers from around 50 snippets for each question.
TriviaQA~\cite{joshi2017triviaqa} collects a set of questions, each along with top-50 web pages including encyclopedic entries and blog articles.
SQuAD-open~\cite{drqa} removes the corresponding articles from each question in SQuAD~\cite{rajpurkar2016squad:}, and is designed for the setting of open-domain question answering.
MS-MARCO~\cite{msmarco} provides 100K questions where each question is matched with 10 web pages. 
And DuReader~\cite{he2017dureader} is a large scale Chinese dataset collected in the same way as MS-MARCO.
Recently, HotpotQA~\cite{yang2018hotpotqa} is collected for multi-hop reasoning among multiple paragraphs, which supports the community to study question answering at a large scale.
All these datasets can advance QA models to deal with more challenging and practical scenarios.

\subsection{Approaches for Open-domain QA}

\paragraph{Pipeline systems}
It is natural to decompose open-domain QA into two stages: retrieving relevant documents by a retriever and extracting the answer from the retrieved documents by a reader.
Chen et al.~\cite{drqa} developed DrQA which first retrieves Wiki documents using bigram hashing and TF-IDF matching, and then extracts answers from top-$K$ articles with a multi-layer RNN RC model.
Seo et al.~\cite{seo2019realtime} introduced the query-agnostic representations of documents to speed up the retriever.
Clark et al.~\cite{documentqa} used a TF-IDF heuristic method to select paragraphs and improve the RC component via a shared normalization to calibrate answer scores among individual paragraphs.
Similarly, Wang et al.~\cite{wang2019multibert} applied shared normalization to the BERT reader when simultaneously dealing with multiple passages for each question.
Ni et al.~\cite{termquery} improved the retriever to attend on key words in a question and reformulated the query before searching for the related evidence.
Yang et al.~\cite{bertserini} proposed BERTserini that integrates the most powerful BERT RC model with the open-source Anserini information retrieval toolkit.
These pipeline systems are straightforward but independent training of different components may face a context inconsistency problem~\cite{re3}.

\paragraph{Joint training models}
In order to address the issue that independent IR components 
do not consider RC components,
a variety of joint training methods have been proposed.
Choi et al.~\cite{choi2017coarse} proposed a coarse-to-fine QA framework aiming at selecting only a few relevant sentences to read. They treated the selected sentence as a latent variable which can be trained jointly, supervised by the final answer using reinforcement learning (RL).
Wang et al.~\cite{wang2018joint} also regarded the candidate document extraction as a latent variable and trained the two-stage process jointly.
Min et al.~\cite{minimal} trained a shared encoder for a sentence selector (IR component) and a reader (RC component).
Nishida et al.~\cite{retrieveandread} used a supervised multi-task learning framework to train the IR component by considering answer spans from the RC component.
Wang et al.~\cite{r3,reranker} presented the $R^3$ system in a retriever-reader-reranker paradigm. The retriever ranks retrieved passages and passes the most relevant passages to the reader. The reader determines the answer candidates and estimates the reward to train the retriever. The reranker reranks the answer candidates with strength-based and coverage-based principles.  
Moreover, Htut et al.~\cite{ranking-function} improved the retriever using relation network~\cite{relationnetwork}, and Wang et al.~\cite{vnet} improved the reranker using a neural model to verify answer candidates from different passages.
In order to capture useful information from full but noisy paragraphs, DS-QA~\cite{dsqa} and HAS-QA~\cite{pang2019has} decomposed the probability of answers into two terms, i.e. the probability of each paragraph by the retriever and the probability of answers given a certain paragraph by the reader.
In such probabilistic formulation, all documents can be considered.
Dehghani et al.~\cite{tracrnet} proposed TraCRNet which adopts the Transformer~\cite{transformer} to efficiently read all candidate documents in case the answers exist in low-ranked or not directly relevant documents.
Recently, Hu et al.~\cite{re3} proposed $RE^3QA$ system which models the retriever, the reader, and the reranker via BERT~\cite{bert} and achieved much better performance.
Joint models improve the consistency of different components, and therefore are more benefical than pipeline systems.

\paragraph{Iterative Frameworks}
Recently, more and more studies have focused on handling more sophisticated situations where single-step retrieval and reasoning may be insufficient.
To fast retrieve and combine information from multiple paragraphs, Das et al.~\cite{interaction} introduced a reader-agnostic architecture where the retriever and the reader iteratively interact with each other. At each step, the query is updated according to the state of the reader, and the reformulated query is used to rerank the pre-cached paragraphs from the retriever.
Peng et al.~\cite{iterativequerygeneration} claimed that not all the relevant context can be obtained in a single retrieval step and proposed GoldEN Retriever to answer open-domain multi-hop questions.
At each step, GoldEN Retriever uses results from previous reasoning hops to generate a new query and retrieve new evidence via an off-the-shelf retriever.
Ding et al.~\cite{ding2019cognitive} designed an iterative framework for multi-hop QA named CogQA, which pays more attention to the reasoning process rather than the retriever. CogQA extracts relevant entities from the current passage to build a cognitive graph, and uses the graph to decide the current answer and next-hop passages.

\subsection{Knowledge in Retrieval-based QA Models}
Our work is also inspired by the research which incorporates knowledge in QA models.
Sun et al.~\cite{fusion} leveraged relevant entities from a KB and relevant text from Wikipedia as external knowledge to answer a question.
Lin et al.~\cite{lin2019kagnet} constructed a schema graph between QA-concept pairs for commonsense reasoning.
In order to retrieve reasoning paths over Wikipedia,  Godbole et al.~\cite{Entity-centric} used entity linking for multi-hop retrieval. 
Asai et al.~\cite{retrievepath} utilized the wikipedia hyperlinks to construct the Wikipedia graph which helps to identify the reasoning path.
Though many efforts have been devoted into designing knowledge-aided reasoning components in QA systems, our work aims at improving the retriever and reranker components through building question-document and document-document graphs with the aid of external knowledge.

\section{Methodology}
In this section, we describe our Knowledge-Aided Question Answering (KAQA) model in detail. Our model follows the retriever-reader-reranker framework~\cite{r3,reranker} but incorporates knowledge into different components.

\begin{figure*}[ht]
  \centering
  \includegraphics[width=\linewidth]{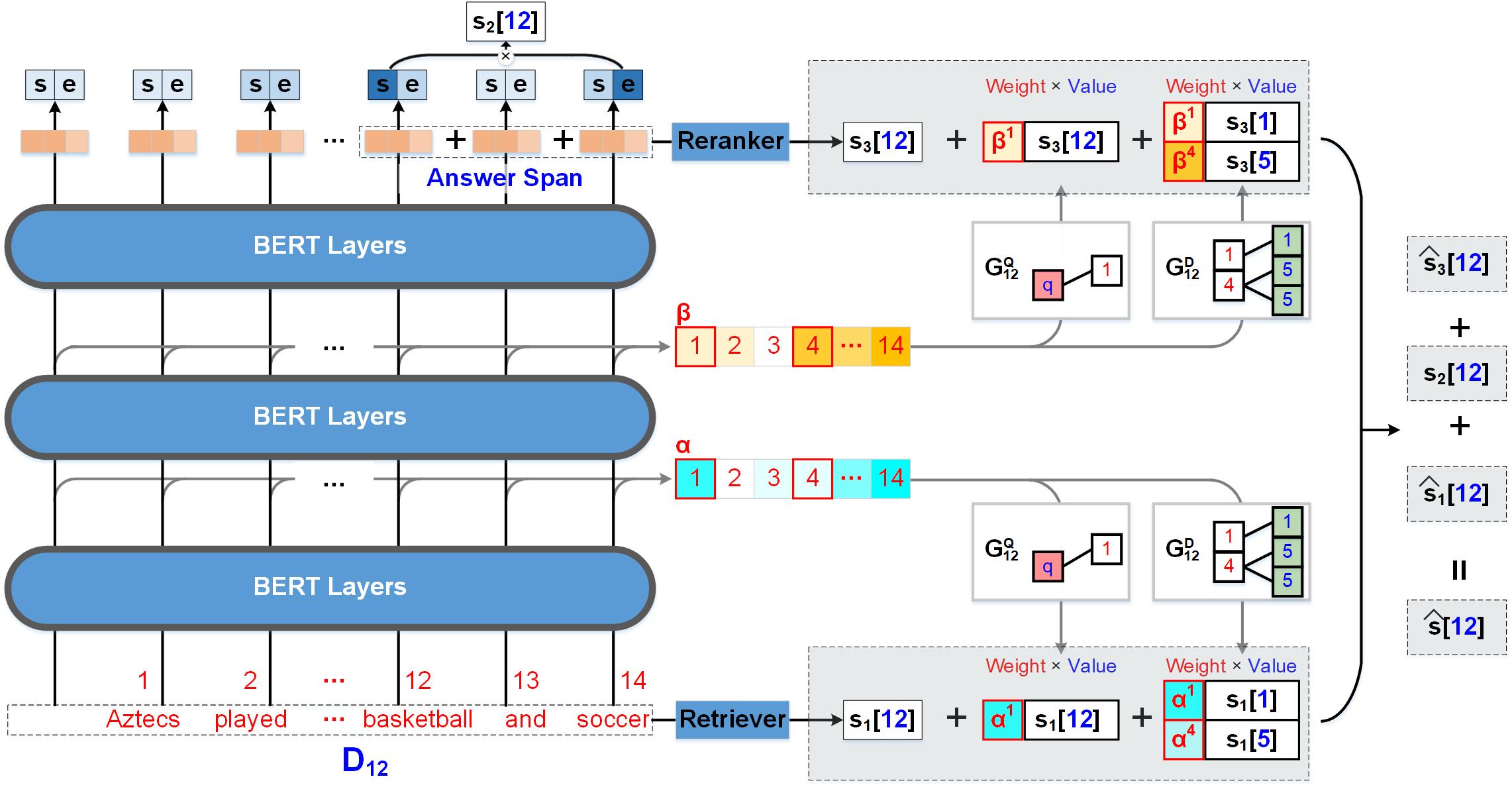}
  \caption{The KAQA framework working on an example. A candidate document $D_{12}$ is first assigned a retrieval score $s_1[12]$ via simple TF-IDF. Then a reader with multiple BERT layers decides a candidate answer span in this document with the largest start/end position probability. And the reranker assigns a reranking score $s_3[12]$ to the candidate answer. Relational knowledge $G^Q_{12}$ and $G^D_{12}$ 
  are utilized to modify the retrieval score and reranking score by leveraging the corresponding scores of other candidates. One internal BERT layer outputs $\alpha$ as the weights for each word (marked in red). The weights of words connected to the question enlarges $s_1[12]$, and the weights of words connected to other documents (document indices marked in blue) are used to integrate the retrieval score of other documents ($s_1[1]$ and $s_1[5]$) into $s_1[12]$. 
  Another internal BERT layer outputs $\beta$ as the similar weights used to modify $s_3[12]$. The final answer is decided by the knowledge-enhanced scores $\hat{s}$.}
  \Description{Model framework.}
  \label{fig:model}
\end{figure*}

Figure~\ref{fig:model} gives an overview of our KAQA model. 
Specifically, each candidate document $D_i$ is first assigned a retrieval score $s_1[i]$ by a simple retriever. 
Then, a reader with multiple BERT layers decides a candidate answer in this document with the largest start/end position probability. 
An MLP reranker assigns a confidence score $s_3$ of the candidate answer afterwards. 
In order to improve the retriever and the reranker, we extract the graph between questions and documents ($G^Q$) and the graph among documents ($G^D$) as the relational knowledge. 
Such knowledge is utilized to refine the retrieval and reranking scores by leveraging the  scores of other candidates.

In what follows, we first introduce the retriever-reader-reranker framework and the knowledge we used, and then we describe each component in turn.

\subsection{Retriever-Reader-Reranker Framework}
Open-domain question answering aims to extract the answer to a given question $Q$ from a large collection of documents $D = \{D_1, D_2, ...,D_N\}$. 
The retriever-reader-reranker framework consists of three components.

\noindent {\bfseries The Retriever ($R_1$)} first scores each candidate document $D_i$ with
\begin{displaymath}
  s_1[i] = R_1(Q,D_i),
\end{displaymath}
where $s_1[i]$ is the score of document $D_i$ given question $Q$.
It then ranks candidate documents according to the scores $s_1[i]~(1\leq i\leq N)$, and returns a few top ranked candidate documents to the reader component.

\noindent {\bfseries The Reader ($R_2$)} extracts a candidate answer from each candidate document independently, which is the same as that in 
single document
RC models. 
For a given document $D_i=[d^1_i,...d^n_i]$, the reader outputs two distributions over all the tokens $d_i$ as the probability of being the start/end position ($P_s$/$P_e$) of the answer respectively:
\begin{displaymath}
  P_s(d_i) = R_2^s(Q,D_i), P_e(d_i) = R_2^e(Q,D_i).
\end{displaymath}
The candidate answer from document $D_i$ is then determined by
\begin{displaymath}
  a_i = [d^{\hat{l}}_i,..,d^{\hat{m}}_i] = \mathop{\arg\max}_{l<m}P_s(d^l_i)P_e(d^m_i)
\end{displaymath}
with a score
\begin{displaymath}
  s_2[i] = P_s(d^l_i)P_e(d^m_i),
\end{displaymath}
where $[d^l_i,..,d^m_i]$ denotes the text span from the $l$-th word to the $m$-th word in document $D_i$. 

\noindent {\bfseries The Reranker ($R_3$)} aggregates the supporting evidence of each candidate answer and re-scores 
each candidate answer $a_i$ as
\begin{displaymath}
  s_3[i] = R_3(Q,a_i).
\end{displaymath}

The final score of the candidate answer from $D_i$ is the weighted sum of scores from the three components:   
\begin{displaymath}
  s[i] = w_1s_1[i] + w_2s_2[i] + w_3s_3[i].
\end{displaymath}
The output answer is then determined by taking the candidate answer with the largest final score from $\{a_1, a_2, \cdots, a_N\}$.

\begin{figure}[t]
  \centering
  \includegraphics[width=\linewidth]{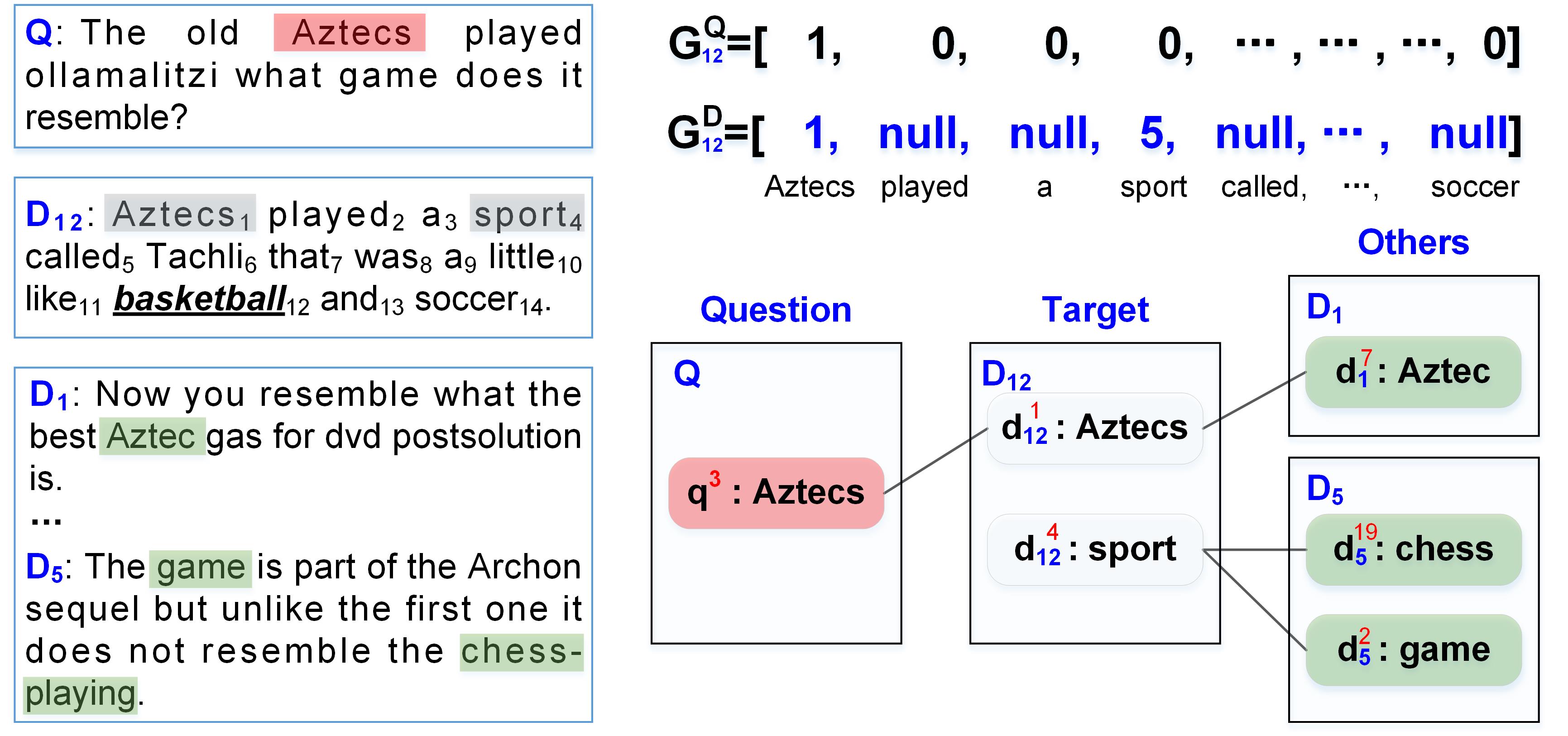}
  \caption{The $D_{12}$-centric subgraph and two corresponding lists $G^Q_{12}$ and $G^D_{12}$.
  Each word in the target document $D_{12}$ is matched with a word in the question or a word in other documents. If the $k$-th word $d_{12}^k$ is related to any word in the question, $G^Q_{12}[k]=1$; otherwise $G^Q_{12}[k]=0$. If $d_{12}^k$ is most related (defined in Eq.~\eqref{eq:GD}) to words in the $j$-th document ($j\neq 12$), $G^D_{12}[k]=j$, otherwise $G^D_{12}[k]=null$. Numbers in blue stand for document indices, and numbers in red stand for word indices in a document.}
  \Description{The subgraph and two lists.}
  \label{fig:graph}
\end{figure}

\subsection{Knowledge-Aided Model}
Our knowledge-aided model improves the original components to be conditioned on external knowledge $K$, i.e. score $s_1[i]$ by the retriever and $s_3[i]$ by the reranker are computed as:
$$s_1[i] = R_1(Q,D_i,K),$$
$$s_3[i] = R_3(Q,a_i,K),$$
where $K$ is involved in $R_1$ and $R_3$.
In our model, the knowledge $K$ includes the question-document graph $G^{Q}$ and the document-document graph $G^{D}$.

For a given question $Q$ with $N$ candidate documents $ \{D_1,...D_N\}$, we extract relational triples with subject and object entities in the question or any candidate document, with the help of external knowledge bases (KBs), such as WordNet.
We build two graphs $G^{Q}$ and $G^{D}$ with these triples,
where each node represents a word in a document or the question, and each edge represents a KB relation.
We also define two lists $G^{Q}_i$ and $G^{D}_i$ for each document $D_i$ to represent the subgraphs of $G^{Q}$ and $G^{D}$ centered at document $D_i$, as shown in Figure~\ref{fig:graph}.

$G^{Q}_i$ is an indicative list which shows whether each word in $D_i$ is connected to any word in the question in ${Q}$:
\begin{equation}
G^Q_i[k]=
\begin{cases}
1& \exists j, r : (d^k_i, r, q^j) \in KBs\\
0& otherwise
\end{cases}
,
\label{eq:GQ}
\end{equation}
where $d^k_i$ represents the $k$-th word in document $D_i$, $q^j$ represents the $j$-th word in the question $Q$, $r$ is a relation or a reversed relation in external knowledge base $K$, and
$G^Q_i[k]=1$ indicates that $d^k_i$ is connected to at least one word in $Q$. 

$G^{D}_i$ is an index list which represents the connection between documents:
\begin{equation}
G^D_i[k]=
\begin{cases}
 \mathop{\arg\max}_{j}|T(j)| &  T(j) =\{ l|(d^k_i, r, d^l_j) \in KBs\}, j\neq i\\
null & otherwise
\end{cases}
\label{eq:GD}
\end{equation}
where $d^k_i$ represents the $k$-th word in document $D_i$ and $d^l_j$ represents the $l$-th word in document $D_j$. 
Since $d^k_i$ can have connections with words in multiple documents, we use $T(j)$ to denote the set of words in $D_j$ that are connected to word $d^k_i$. 
And therefore $|T(j)|$ serves as the strength (the number of relevant words) of the connection between $D_j$ w.r.t. word $d^k_i$. 
$G^{D}_i[k]=j$ stands for that document $D_j$ has the most words connected to $d^k_i$ than other documents . $G^{D}_i[k]=null$ means $d^k_i$ is removed from $G^D$  or $d^k_i$ has no connections to all other documents (see Section~\ref{sec:preprocess}).

\subsection{Retriever}
The retriever is responsible for identifying whether a candidate document contains the answer or not. 
First, the semantic similarity of the question and a candidate document is measured by the cosine value of their TF-IDF vectors:
\begin{displaymath}
 s_1[i] = cos(TF-IDF(Q), TF-IDF(D_i)).
\end{displaymath}
Thereby, we obtain independent retrieval scores for each candidate document:
\begin{displaymath}
 s_1 = \{s_1[1], s_1[2],...s_1[N]\}.
\end{displaymath}

To utilize knowledge $G^Q$ and $G^D$,
we combine the simple retrieval scores with another two knowledge-aided terms as:
\begin{equation}
    s^Q_1[i] = \sum_{k} G^Q_i[k] \alpha^k_i s_1[i],
    \notag
\end{equation}
\begin{equation}
    s^D_1[i] = \sum_{k}\alpha^k_is_1[G^D_i[k]],
    \notag
\end{equation}
\begin{equation}
    \hat{s}_1[i] = s_1[i] + \omega^Qs^Q_1[i] + \omega^Ds^D_1[i],
\label{eq:retriever-score}
\end{equation}
where $\omega^Q$ and $\omega^D$ are the weights to balance the original score and the knowledge-aided terms. $\alpha^k_i$ is the weight assigned to word $d^k_i$. We use the self-attention scores~\cite{transformer} computed among the hidden states from an internal BERT layer as the weights in the knowledge-aided retrieval score. 

In Eq. \eqref{eq:retriever-score}, the retrieval score for $D_i$ is strengthened by the question-document connections $G^Q_i$ via $s^Q_1[i]$ (named as q-link term), and the document-document connections $G^D_i$ via $s^D_1[i]$ (named as d-link term).
The $s^Q_1[i]$ term pays attention to the words connected to the question. If $d^k_i$ is connected to the question, i.e. $G^Q_i[k]=1$, $s_1[i]$ will be enhanced by its weight $\alpha^k_i$. Otherwise, if there is no word connected to the question, the $s^Q_1[i]$ term will be $0$ and there is no additional score from $G^Q$.
The $s^D_1[i]$ term emphasizes the information of other relevant documents. If $d^k_i$ is connected to the $j$-th document, i.e. $G^D_i[k]=j$, $D_j$'s retrieval score $s_1[j]$ will be added to $s_1[i]$ with weight $\alpha^k_i$. Otherwise, if $D_i$ has no connection with other documents, the retrieval score will not be affected by the retrieval scores of other documents.

In the above formulation, we use $\alpha^k_i$ to weigh the importance of each word $d^k_i$ in $D_i$. This is implemented by applying self-attention on the output of an intermediate layer ($L_{\alpha}$), as follows:
\begin{displaymath}
\alpha_i = SelfAttn(H_{L_{\alpha}}) \in \mathcal{R}^{|D_i|},
\end{displaymath}
where $\alpha_i$ is normalized only over tokens in $D_i$ ($|\alpha_i|$ = $|D_i|$) but is also conditioned on the question $Q$ since the model input 
includes $Q$. 
More specifically, the input token sequence to the BERT is as follows: 
\begin{displaymath}
X = [CLS],q^1,...q^{|Q|},[SEP],d^1_i,...d^{|D|}_i,[SEP],
\end{displaymath}
\begin{displaymath}
H_0 = Embed(X).  
\end{displaymath}
And the output of each layer is $H_l$, given by
\begin{displaymath}
H_l = BERTLayer(H_{l-1}), 0 \textless l \leq L.
\end{displaymath}

Note that we set $L_{\alpha}=3$ in our experiments, which is the same as $RE^3$~\cite{re3}. 
The major reason is due to the computational efficiency concern: extracting representations from shallow layers is more efficient than from higher layers, without encoding with all the
BERT 
layers when dealing with large document collections.



\subsection{Reader}
The reader is responsible for identifying the start position and the end position of the answer span in a candidate document $a_i = [d^l_i,..,d^m_i]$. 
Following state-of-the-art RC models, we adopt the BERT-based model~\cite{bert} as our reader.
In this reader, each token $d^k_i$ in document $D_i$ is represented by the concatenation of its token embedding (word embedding), segment embedding and position embedding in the input layer as $H^0$. 
Next, a series of pre-trained 
BERT layers 
are used to encode the input embeddings into a sequence of contextualized vectors as:
\begin{displaymath}
H_0 = Embed(X),
\end{displaymath}
\begin{displaymath}
H_l = BERTLayer(H_{l-1}), 0 \textless l \leq L.
\end{displaymath}
Finally, two different MLP layers transform the hidden vectors from the top layer into two distributions as the possibility of being the start and the end positions respectively:
\begin{displaymath}
P_s(d^k_i)= MLP_s(h_{L}^k),
\end{displaymath}
\begin{displaymath}
P_e(d^k_i)= MLP_e(h_{L}^k),
\end{displaymath}
where $h_{L}^k$ represents the hidden vector on the $k$-th position in the $L$-th BERT layer ($H_L$). $P_s(d^k_i)$ is the probability of the $k$-th word $d^k_i$ being the start of the candidate answer in $D_i$, and $P_e(d^k_i)$ is the probability of word $d^k_i$ being the end of answer span.
The candidate answer is determined by the product of the two distributions:
\begin{displaymath}
  a_i = [d^{\hat{l}}_i,..,d^{\hat{m}}_i] = \mathop{\arg\max}_{l<m}P_s(d^l_i)P_e(d^m_i).
\end{displaymath}

\subsection{Reranker}
The reranker aggregates evidence supporting for each candidate answer and re-score the confidence of each candidate answer.
Each candidate answer is a text span in a single input document, and the reranker first aggregates the BERT representations of words in span $a_i = [d^{\hat{l}}_i,..,d^{\hat{m}}_i] $ to get the summarized answer representation $h_a$ as:
\begin{displaymath}
  \gamma^k = SelfAttn(h_{L}^k),
\end{displaymath}
\begin{displaymath}
  h_{a_i} = \sum_{k=l}^m \gamma^k h_{L}^k,
\end{displaymath}
where 
\begin{math}
\gamma^k
\end{math}
is computed by the self-attention~\cite{transformer} on the hidden states of the last BERT layer $h_L^k$.
Then an MLP outputs the confidence score:
\begin{displaymath}
  s_3[i] = MLP_a(h_{a_i}).
\end{displaymath}
And we obtain the confidence scores of all the candidate answers as:
\begin{displaymath}
 s_3 = \{s_3[1], s_3[2],...s_3[N]\}.
\end{displaymath}
But independent reranking scores may be sub-optimal. 
To utilize knowledge $G^D$ and $G^Q$, we also combine the reranking score with two knowledge-aided terms as:
\begin{equation}
    h^Q_{a_i} = \sum_{k=l}^m G^Q_i[k] \beta^k_i h_{L}^k,
    \notag
\end{equation}
\begin{equation}
    s^Q_3[i] = MLP_a(h^Q_{a_i}),
    \notag
\end{equation}
\begin{equation}
    s^D_3[i] = \sum_{k}\beta^k_is_3[G^D_i[k]],
    \notag
\end{equation}
\begin{equation}
    \hat{s}_3[i] = s_3[i] + \omega^Qs^Q_3[i] + \omega^Ds^D_3[i],
\label{eq:reranker-score}
\end{equation}
where $\beta_i$ is the self-attention assigned to words in $D_i$ obtained similarly to $\alpha_i$ but with hidden states from a deeper layer:
\begin{displaymath}
\beta_i = SelfAttn(H_{L_{\beta}}) \in \mathcal{R}^{|D_i|}.
\end{displaymath}
In the knowledge-aided reranker, we only consider the words in the answer span $a_i$ rather than all the words in $D_i$.
In Eq. \eqref{eq:reranker-score}, the reranking score is enhanced by $G^Q$ and $G^D$.
The $s^Q_3[i]$ term emphasizes the words connected to the question. If $d^k_i$ is connected to the question, i.e. $G^Q_i[k]=1$, $s_3[i]$ will be enhanced by its weight $\beta^k_i$. Otherwise, if no word is connected to the question, the score will not be enhanced.
The $s^D_3[i]$ term emphasizes the information of other relevant documents. If $d^k_i$ is connected to the $j$-th document, i.e. $G^D_i[k]=j$, $s_3[j]$ will be added to $s_3[i]$ with the weight $\beta^k_i$. Otherwise, if $D_i$ has no connection with other documents, it can not be supported by others' retrieval scores.

Note that we set $L_{\beta}=11$ in our experiments because the vectors from the second last layer are verified effective when using as features in other tasks~\cite{bert}.

\subsection{Inference and Optimization}
The final answer is determined the overall score which sums the modified retrieval and reranking scores as well as the original reader score, as follows:       
\begin{equation}
  \hat{s}[i] = w_1\hat{s}_1[i] + w_2s_2[i] + w_3\hat{s}_3[i].
  \label{eq:final-score}
\end{equation} 
The model is optimized jointly by the supervision applied to each component as
\begin{equation}
  \mathcal{L} =  \mathcal{L}_1 + \mathcal{L}_2  + \mathcal{L}_3.
  \label{eq:final-loss}
\end{equation}

For the retriever, a binary score $s^*_1[i]$ indicating whether each document includes the golden answer supervises the retrieval score $\hat{s}_1[i]$ as
\begin{displaymath}
  \mathcal{L}_1 = \sum_i \parallel sigmoid(\hat{s}_1[i])  - s^*_1[i]\parallel_2.
\end{displaymath}

For the BERT-based reader which outputs the $\alpha$ and $\beta$ for the knowledge-aided retriever and reranker, the golden answer span $a^*$ supervises the model via cross-entropy loss:
\begin{displaymath}
 \mathcal{L}_2 = -\log P_s(a^*) - \log P_e(a^*).
\end{displaymath}

Motivated by $RE^3$~\cite{re3}, the reranking score $\hat{s}_3$ can also be supervised by a hard label $s^{hard}_3$ and a soft label $s^{soft}_3$ as follows: 
\begin{displaymath}
 \hat{s}^{hard}_3[i] \xleftarrow{} softmax(\hat{s}_3[i]),
\end{displaymath}
\begin{displaymath}
 \hat{s}^{soft}_3[i] \xleftarrow{} sigmoid(\hat{s}_3[i]),
\end{displaymath}
\begin{displaymath}
 \mathcal{L}_3 = \sum|| \hat{s}^{soft}_3 - s^{soft}_3||_2 - \sum s^{hard}_3\log(\hat{s}^{hard}_3),
\end{displaymath}
where $s^{soft}_3[i]$ is the F1 score of the candidate answer $a_i$ compared to the golden answer $a^*$, and $s^{hard}_3[i]$ is defined as the EM score of $a_i$. The softmax operation converts the reranking scores of all candidate answers to the question into probabilities, and the label $s^{hard}_3$ has value 1 on only the ground truth answer (0-1 distribution).

\section{Experiment}
\subsection{Datasets and Preprocessing}
\label{sec:preprocess}
We have conducted experiments on three public open-domain question answering datasets: SQuAD-open~\cite{drqa}, Quasar-T~\cite{dhingra2017quasar} and TriviaQA~\cite{joshi2017triviaqa}. For each dataset, we train the models on the training set and evaluate the models on the development set. The statistics of these datasets are shown in Table~\ref{tab:stat}.

{\bfseries SQuAD-open} is the open-domain version of SQuAD (a benchmark for single document reading comprehension). In SQuAD-open, the evidence corpus for each question is the entire Wikipedia rather than a specific paragraph of a Wikipedia article. 
For each question, we sifted out top 10 documents as the input candidates using TF-IDF similarity scores.

{\bfseries Quasar-T} consists of trivia questions with long and short contexts extracted from 
ClueWeb09 corpus using the ``Lucene index''. For each question, 100 unique sentence-level documents were collected. The answers to the questions were obtained from various internet sources and may not exist in any candidate document. 
Since the answer coverage of long contexts and short contexts are almost the same, we only use the short contexts as the candidate documents for the computational efficiency. Only the instances where the answer can be extracted from the documents are retained. 

{\bfseries TriviaQA} is a large dataset for the reading comprehension task and the open-domain QA task. We focus on the open-domain QA subset, i.e. TriviaQA-unfiltered-web, which gathers question-answer pairs from 14 trivia and quiz-league websites, and collects the top 50 search results returned from the Bing Web search API as the relevant documents for each question. The search output includes a diverse set of documents such as blog articles, news articles, and encyclopedic entries. Only the top 10 results keep entire Web pages or PDF, and the others only contain URLs, titles and descriptions.
In our experiments, we only keep the instances where the answer can be extracted from the documents.

Due to the lack of answer annotations in Quasar-T and TriviaQA datasets,  the golden document is automatically labeled as the first document (in the ranked order) where the golden answer can be exactly matched with a text span in the document, and the position of the span is used to supervise our model.
Instances with more than 8,000 tokens (words in the question and all the candidate documents) are removed.
\begin{table}[ht]
  \caption{The statistics of three open-domain QA datasets.}
  \label{tab:stat}
  \adjustbox{max width=.45\textwidth}{
      \begin{tabular}{cccc}
        \toprule
        Datasets & \#Training examples & \#Development examples & \#Avg. docs \\
        \midrule
        SQuAD-open & 87,599 & 10,570 & 10\\
        Quasar-T & 25,465 & 2,068 & 100\\
        TriviaQA & 68,001 & 8,768 & 50 \\
      \bottomrule
    \end{tabular}
    }   
\end{table}

To construct $G^Q$ and $G^D$, we first extract noun phrases in the questions and the documents with NLTK\footnote{\url{http://www.nltk.org}} and spaCy\footnote{\url{https://spacy.io}} toolkits.
Then, we pair one noun phrase from a document with another one from the question or another document. We check whether each pair is a valid triple\footnote{Semantic different relations, such as ``/r/Antonym", are excluded.} defined in the knowledge bases. In our experiments, we use WordNet \citep{wordnet}, Freebase\citep{freebase}, and ConceptNet\citep{liu2004conceptnet} as external knowledge bases.
If one noun phrase in the question is connected to more than $T_1$ documents,  we removed all the edges connected to this phrase in $G^Q$ since such common nodes provide little information to distinguish candidate documents.
The threshold $T_1$ is set to 10 for Quasar-T and TriviaQA and 5 for SQuAD-open, since each question in SQuAD-open is only paired with 10 candidate documents.
We also limit the number of nodes in $G^D$ to prevent the d-link term ($s^D_1$) from leaning towards long documents. For each document, we keep at most $T_2$ words (nodes) which are connected to another document in $G^D$. Such words are ranked according to their inverse document frequency (IDF). Common words with small IDF are removed from the $G^D$ if a document has too many outgoing edges. 
$T_2$ is set to 10 for Quasar-T and TriviaQA, and is set to 30 for SQuAD-open, since the documents in SQuAD-open are much longer.

\subsection{Experimental Settings}
We initialize our model using the uncased version of BERT-base~\cite{bert}.
We first fine-tune the reader (with only $\mathcal{L}_2$) for 1 epoch, and then fine-tune the whole model (with $\mathcal{L}$) for another 2 epoches.
We use Adam optimizer with a learning rate of $3\times 10^{-5}$ and the batch size is set to 32.

The pre-trained BERT reader has $L=12$ layers.
The shallower internal layer used to compute $\alpha$ for the retriever is $L_{\alpha}=3$, which is the same as $RE^3$~\cite{re3}. And the deeper internal layer used to compute $\beta$ for the reranker is $L_{\beta}=11$ since most approaches based on BERT suggest to use the vectors from the second last layer~\cite{bert}.

The weights to incorporate $G^Q$ and $G^D$ in the retriever (Eq. \eqref{eq:retriever-score}) and the reranker (Eq. \eqref{eq:reranker-score}) are $\omega^Q=0.5$, and $\omega^D=0.5$.
The weights to balance scores from different components in the final score (Eq. \eqref{eq:final-score}) are searched from $w_{1/2/3} \in \{0.2,0.5,1.0\}$.
And all the loss terms in Eq. \eqref{eq:final-loss} are normalized to the same scale.

\subsection{Preliminary Experiments}
\label{sec:pre_exp}
Most open-domain QA systems only provide the reader with top-5 retrieved documents for answer extraction. We evaluate how well the retriever ranks the golden documents that contain correct answers.
Results in Table~\ref{tab:freq} show that TriviaQA has good retrieval results where more than $80\%$ golden documents can be found in top 5 positions. However, the cases for SQuAD-open and Quasar-T are quite unsatisfactory. There are about 50\% chances that the golden documents are not passed to the reader, where in such cases wrong answers will be produced inevitably.
\begin{table}[ht]
  \caption{Ranking performance evaluated on the development sets of different open-domain QA datasets. SQuAD-open is ranked by TF-IDF similarity. Quasar-T and TriviaQA are already ranked in search order by the original datasets.}
  \label{tab:freq}
  \begin{tabular}{cccc}
    \toprule
    Datasets & P@3 & P@5 & P@10\\
    \midrule
    SQuAD-open & 52.9 & 59.8 & 67.7 \\
    Quasar-T & 39.2 & 48.0 & 56.9 \\
    TriviaQA & 72.6 & 80.9 & 89.5 \\
  \bottomrule
\end{tabular}
\end{table}

Furthermore, we conduct experiments where golden documents are enforced to pass to the reader to verify the effect with a high-quality retriever.
We evaluate the performance of the state-of-the-art baseline model $RE^3$~\cite{re3} under different settings.
In the original setting, the reader in $RE^3$ receives the top-N documents ranked by the retriever. If the retriever assigns a low score to the golden document, the reader can not extract the correct answer.
In the ``+ Golden Doc'' setting, if a golden document is excluded from top-N candidates, we manually replace the $N$-th candidate with the golden document.
Results in Table~\ref{tab:upperbound} show that, even the reader and reranker are still imperfect, the performance has been boosted substantially when the golden documents can be correctly retrieved. 
Though the retrieval performance is quite high on TriviaQA (P@5/ P@10 is about 80\%/90\% respectively, as shown in Table \ref{tab:freq}), better retrievers can still improve the performance of the open-domain QA system remarkably (from 69.8\% to 77.4\% in F1).

We further evaluate the ``+ All Answer'' setting where all candidate answers extracted by the reader rather than only the most possible answer are compared with the ground truth answer.
We evaluate the maximum EM and F1 values over all candidate answers. This experiment indicates the upper bound of the performance if the reranker is perfect.
Results in the last row of Table~\ref{tab:upperbound} show that, for TriviaQA, nearly $15\%$ F1 decrease is caused by the reranker when wrongly assigning a lower confidence score to the correct candidate answers which have already been extracted by the reader. And the decrease is even larger for Quasar-T and SQuAD-open.

These experiments show that though powerful BERT based models have achieved strong performance on single document machine comprehension, there is still much room for improvement in open-domain QA systems due to the limitation of the retriever and the reranker.
Results show that the performance of question answering can be substantially improved if the retriever and the reranker have better performance.

\begin{table}[h]
  \caption{QA performance of $RE^3$ model with the perfect retriever (+Golden Doc) and perfect reranker (+All Answer).}
  \label{tab:upperbound}
  \begin{tabular}{l|cccccc}
    \toprule
    \multirow{2}{*}{Methods} &\multicolumn{2}{c}{SQuAD-open} & \multicolumn{2}{c}{Quasar-T} & \multicolumn{2}{c}{TriviaQA}\\
    & EM & F1 & EM & F1 & EM & F1\\
    \midrule
    $RE^3$ & 40.1 & 48.4 & 55.8 & 60.8 & 64.1 & 69.8 \\
    + Golden Doc & 68.9 & 75.5  & 59.7 & 64.7 & 72.1 & 77.4 \\
    + All Answer & 55.2 & 65.0 & 82.8 & 86.4 & 78.4 & 84.7\\
  \bottomrule
\end{tabular}
\end{table}

\subsection{Overall Performance}

\begin{table*}[ht]
  \caption{Exact Match (EM) and F1 scores of different models on SQuAD-open, Quasar-T and TriviaQA development sets. 
  For previous models: $^\dagger$ indicates that the model is based on BERT-large while our model uses BERT-base (a smaller version of BERT); $*$ indicates that the results are obtained by ourselves using open-source codes; other values are directly copied from the original papers; and ``-'' indicates that the original papers do not provide the corresponding scores. }
  \label{tab:main}
  \begin{tabular}{lcccccc}
    \toprule
    \multirow{2}{*}{Models} &\multicolumn{2}{c}{SQuAD-open} & \multicolumn{2}{c}{Quasar-T} & \multicolumn{2}{c}{TriviaQA-unfiltered}\\
    & EM & F1 & EM & F1 & EM & F1\\
    \midrule
    DrQA~\cite{drqa} & 27.1 & - & - & - & 32.3 & 38.3\\
    $R^3$~\cite{r3} & 29.1 & 37.5 & 34.2 & 40.9 & 47.3 & 53.7\\
    ReRanker~\cite{reranker} & - & - & 42.3 & 49.6 & 50.6 & 57.3\\
    DS-QA~\cite{dsqa} & 28.7 & 36.6 & 42.2 & 49.3 & 48.7 & 56.3\\
    Shared-Norm~\cite{documentqa} & - & - & 38.6 & 45.4 & 61.3 & 67.2\\
    Retrieve-and-Read~\cite{retrieveandread} & 32.7 & 39.8 & - & - & - & -\\
    Extraction + Selection (Joint Training)~\cite{wang2018joint} & - & - & 45.9 & 53.9 & - & - \\
    
    MINIMAL~\cite{minimal}  & 34.7 & 42.5 & & & - & -\\
     TraCRNet~\cite{tracrnet} & - & - & 43.2 & 54.0 & - & -\\
    HAS-QA~\cite{pang2019has} & - & - & 43.2 &  48.9& 63.6 & 68.9\\
    Multi-Step~\cite{interaction} & 31.9 & 39.2 & 39.5 & 46.7 & 55.9 & 61.7 \\
    BERTserini~\cite{bertserini} & 38.6 & 46.1 & - & - & - & -\\
    Multi-Passage BERT~\cite{wang2019multibert} &  53.0$^\dagger$ & 60.9$^\dagger$ & 51.1$^\dagger$ & 59.1$^\dagger$ & 63.7$^\dagger$ & 69.2$^\dagger$ \\
    $RE^3$~\cite{re3} & 40.1 & 48.4 & 55.8$*$ & 60.8$*$ & 64.1& 69.8\\
    \midrule
    ours & \bf 43.6 & \bf 53.4 & \bf 57.3 & \bf 62.2 & \bf 66.6 & \bf 72.2\\
    \bottomrule 
  \end{tabular}
\end{table*}

We verify the performance of our model on the development sets of the datasets. Table~\ref{tab:main} shows the Exact Match (EM) scores and F1 scores of our model as well as the scores of previous models.
We can observe that:

\begin{itemize}

\item Except the Multi-Passage BERT~\cite{wang2019multibert} which adopts BERT-large, our model achieved the best performance on SQuAD-open.
Our model is remarkably better than $RE^3$~\cite{re3} which also adopts BERT-base,
demonstrating the advantages of our proposed retriever and reranker.

\item Our model outperforms all the other methods on Quasar-T. It is reasonable that our model has an F1 performance of 62.2\% on this dataset since the room for improvement implied by using golden documents as illustrated in Table~\ref{tab:upperbound} is only $3.9\%$ F1 score (from 60.8\% to 64.7\%) and much smaller than those on the other two datasets. 
Larger performance gains may require more powerful reasoning models (namely the reader component).

\item Our model also outperforms all baselines on TriviaQA-unfiltered with a relative smaller improvement because the retriever on this dataset already performs well (80.9\% of P@5) as shown in Table~\ref{tab:freq}.
\end{itemize}
These results demonstrate the advantages of our proposed model.

\subsection{Analysis on the Retriever}
\begin{figure}[h]
  \centering
  \includegraphics[width=\linewidth]{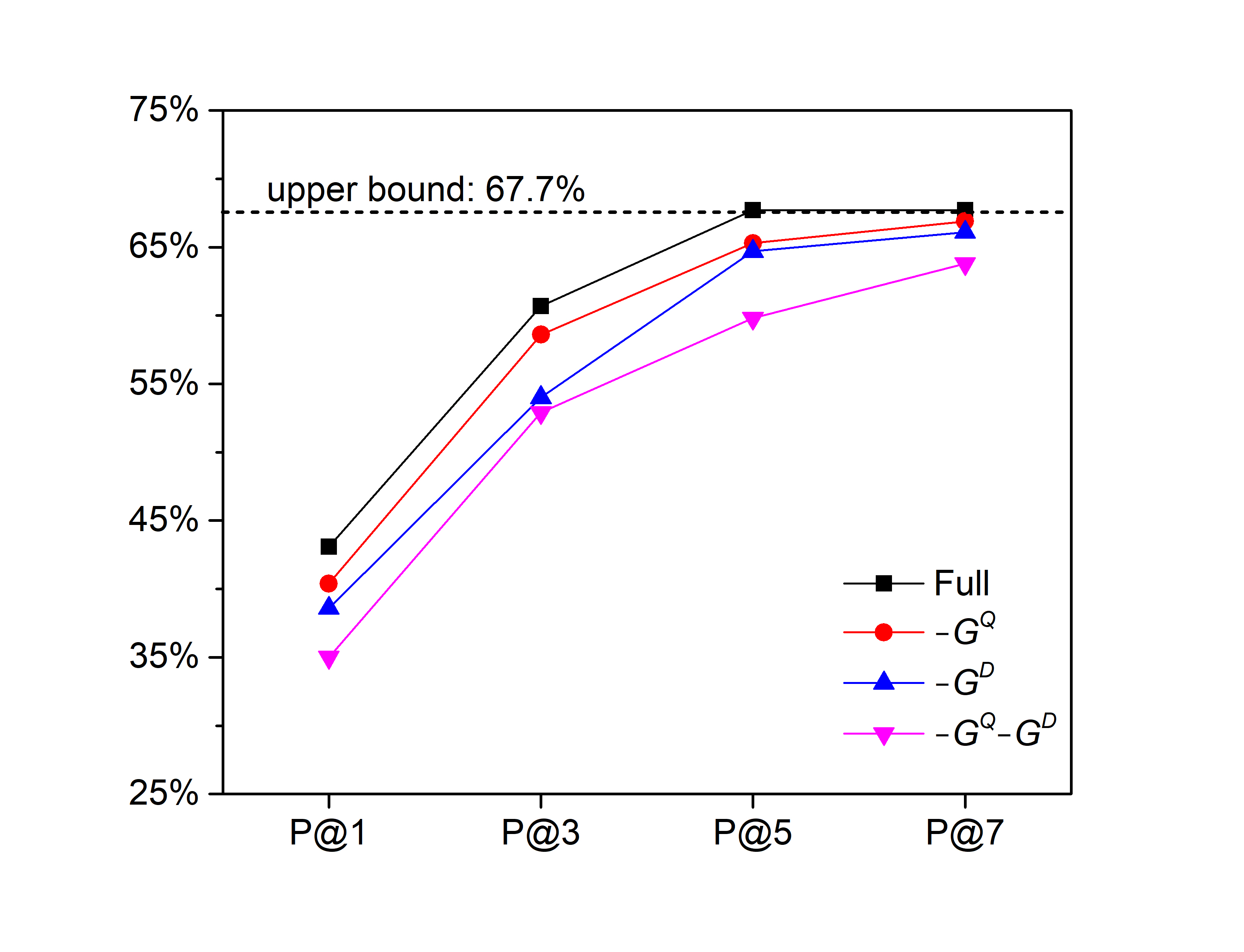}
  \caption{Retrieval performance of different models on the SQuAD-open development set.}
  \Description{Hit k retrieval.}
  \label{fig:hit_k}
\end{figure}

As our model incorporates different types of knowledge,
we conduct ablation tests to investigate the effect of using the question-document graph ($G^Q$) and the document-document graph ($G^D$) on the retriever component.
We demonstrate here how the knowledge ($G^Q$ and $G^D$) affects the rank of golden documents on the development set of SQuAD-open in Figure~\ref{fig:hit_k} and report the final F1 scores on the three datasets in Table~\ref{tab:retriever-abla}.

First, the retrieval performance increases consistently when incorporating $G^Q$ and $G^D$.
The full model performs the best and even approaches the upper bound\footnote{Since there are only 10 documents per question with $67.7\%$ recall in our experiments on SQuAD-open.} in our experiments,
consistently demonstrating that both $G^Q$ and $G^D$ in the retriever benefit the performance of open-domain QA systems.

Second, the contribution of $G^Q$ is larger than that of $G^D$ on SQuAD-open (``removing $G^Q$'': $\downarrow 2.9\%$ F1 v.s. ``removing $G^D$'': $\downarrow 1.3\%$ F1). But the situation is reversed on Quasar-T (``removing $G^Q$'': $\downarrow 0.5\%$ F1 v.s. ``removing $G^D$'': $\downarrow 0.9\%$ F1). This difference may be explained by the fact that the links between Wikipedia articles in SQuAD-open are not as dense as those between the Web pages in Quasar-T.

\begin{table}[h]
  \caption{Ablation tests when removing different graphs from the retriever. F1 scores are reported.}
  \label{tab:retriever-abla}
  \begin{tabular}{l|ccc}
    \toprule
    Retriever & SQuAD-open & Quasar-T & TriviaQA  \\
    \midrule
    Full model & 53.4 & 62.2 & 72.2 \\
    - $G^Q$ & 50.5 & 61.7 &71.6\\
    - $G^D$ & 52.1 & 61.3 &71.9 \\
    - $G^Q$ - $G^D$ & 49.8 & 61.0 &71.2\\
  \bottomrule
\end{tabular}
\end{table}

\subsection{Analysis on the Reranker}
We further conduct ablation tests to investigate the effect of question-document graph ($G^Q$) and document-document graph ($G^D$) on the reranker component.

\begin{figure}[h]
  \centering
  \includegraphics[width=\linewidth]{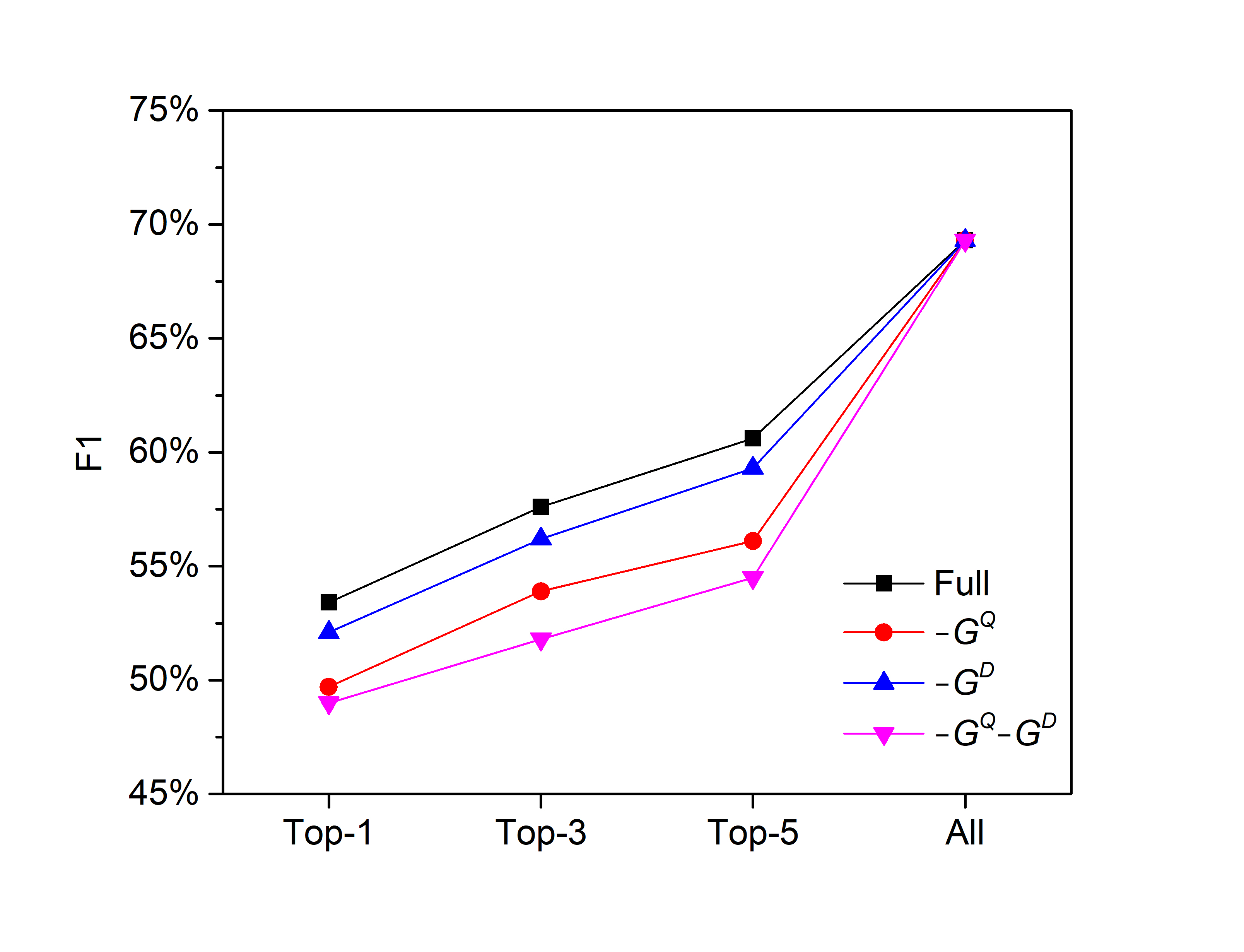}
  \caption{Maximum F1 scores over top-N documents and all candidate answers from different rerankers on the SQuAD-open development set.}
  \Description{Top k F1.}
  
  \label{fig:top_k}
\end{figure}

We first evaluate the improvement of answer reranking from different rerankers on the SQuAD-open development set. We keep the retriever and the reader unchanged and thus obtain the same candidate answers for each question. Then different rerankers assign a confidence score to each candidate answer. We compare the maximum F1 scores over the top-1, top-3, top-5 and all candidate answers. 
As illustrated in Figure~\ref{fig:top_k}, the reranker can be improved to assign larger confidence scores to correct answers with higher F1 scores when incorporating $G^Q$ and $G^D$.

We then evaluate the final performance on the three datasets. 
The results in Table~\ref{tab:retriever-abla}
consistently show that the removal of either $G^Q$ or $G^D$ can degrade the model performance, and thus both $G^Q$ and $G^D$ in the reranker contribute to the performance of open-domain QA systems.

For one thing, we observe that the performance drop is smaller when removing $G^Q$ and $G^D$ in the reranker than that when removing them in the retriever, indicating that the two relational graphs are more beneficial to the retriever.
For another thing, different from the observations in the analysis on the retriever, the contribution of $G^Q$ to the reranker is smaller than that of $G^D$ on all the datasets.
Since the input of the reranker is only the candidate answer spans, which probably have few words connected to the question, the q-link term ($s^Q_3$) is probably zero. 
However, the correct answer is more likely to occur in multiple documents~\cite{reranker}, and thus aggregating information from document-document graphs ($G^D$) may benefit answer reranking more remarkably.

\begin{table}[h]
  \caption{Ablation tests when different information is removed from the reranker. F1 scores are reported.}
  \label{tab:reranker-abla}
  \begin{tabular}{l|ccc}
    \toprule
    Reranker & SQuAD-open & QuasarT & TriviaQA \\
    \midrule
    Full model & 53.4 & 62.2 & 72.2\\
    - $G^Q$ & 53.0 & 61.9 & 71.8\\
    - $G^D$ & 51.2 & 61.5 &71.3 \\
    - $G^Q$ - $G^D$ & 50.3 & 61.2 & 70.9\\
  \bottomrule
\end{tabular}
\end{table}

\section{Conclusion}
This paper investigates how the performance of open-domain question answering can be improved through enhancing document retrieval and answer reranking. The central idea is to consider both question-document and document-document relationships in the document retriever and the answer reranker. More specifically, with the aid of external knowledge resources, we first construct question-document graphs and document-document graphs using knowledge triples, and then encode such relational knowledge in the document retrieval and answer ranking components. 

We evaluated our model on several open-domain question answering datasets including SQuAD-open, Quasar-T and TriviaQA-unfiltered. We observed that our method can boost the overall performance of open-domain question answering consistently on these datasets. Extensive experiments show that modeling the question-document and document-document relationships can contribute to the improvement consistently.

Though our method is simple and effective, we plan to use more sophisticated models such as graph convolutional networks to incorporate such relational knowledge into open-domain QA systems. We leave this as future work.


\bibliographystyle{acm}
\bibliography{reference}

\begin{thebibliography}{10}

\bibitem{retrievepath}
{\sc Asai, A., Hashimoto, K., Hajishirzi, H., Socher, R., and Xiong, C.}
\newblock Learning to retrieve reasoning paths over wikipedia graph for
  question answering.
\newblock {\em arXiv preprint arXiv:1911.10470\/} (2019).

\bibitem{freebase}
{\sc Bollacker, K., Evans, C., Paritosh, P., Sturge, T., and Taylor, J.}
\newblock Freebase: a collaboratively created graph database for structuring
  human knowledge.
\newblock In {\em Proceedings of the 2008 ACM SIGMOD international conference
  on Management of data\/} (2008), AcM, pp.~1247--1250.

\bibitem{drqa}
{\sc Chen, D., Fisch, A., Weston, J., and Bordes, A.}
\newblock Reading wikipedia to answer open-domain questions.
\newblock {\em In Proceedings of the 55th An- nual Meeting of the Association
  for Computational Linguistics, ACL\/} (2017).

\bibitem{choi2017coarse}
{\sc Choi, E., Hewlett, D., Uszkoreit, J., Polosukhin, I., Lacoste, A., and
  Berant, J.}
\newblock Coarse-to-fine question answering for long documents.
\newblock In {\em Proceedings of the 55th Annual Meeting of the Association for
  Computational Linguistics (Volume 1: Long Papers)\/} (2017), pp.~209--220.

\bibitem{documentqa}
{\sc Clark, C., and Gardner, M.}
\newblock Simple and effective multi-paragraph reading comprehension.
\newblock {\em arXiv preprint arXiv:1710.10723\/} (2017).

\bibitem{interaction}
{\sc Das, R., Dhuliawala, S., Zaheer, M., and McCallum, A.}
\newblock Multi-step retriever-reader interaction for scalable open-domain
  question answering.
\newblock {\em arXiv preprint arXiv:1905.05733\/} (2019).

\bibitem{tracrnet}
{\sc Dehghani, M., Azarbonyad, H., Kamps, J., and de~Rijke, M.}
\newblock Learning to transform, combine, and reason in open-domain question
  answering.
\newblock In {\em WSDM\/} (2019), pp.~681--689.

\bibitem{bert}
{\sc Devlin, J., Chang, M.-W., Lee, K., and Toutanova, K.}
\newblock Bert: Pre-training of deep bidirectional transformers for language
  understanding.
\newblock {\em arXiv preprint arXiv:1810.04805\/} (2018).

\bibitem{dhingra2017quasar}
{\sc Dhingra, B., Mazaitis, K., and Cohen, W.~W.}
\newblock Quasar: Datasets for question answering by search and reading.
\newblock {\em arXiv preprint arXiv:1707.03904\/} (2017).

\bibitem{ding2019cognitive}
{\sc Ding, M., Zhou, C., Chen, Q., Yang, H., and Tang, J.}
\newblock Cognitive graph for multi-hop reading comprehension at scale.
\newblock {\em ACL\/} (2019).

\bibitem{dunn2017searchqa}
{\sc Dunn, M., Sagun, L., Higgins, M., Guney, V.~U., Cirik, V., and Cho, K.}
\newblock Searchqa: A new q\&a dataset augmented with context from a search
  engine.
\newblock {\em arXiv preprint arXiv:1704.05179\/} (2017).

\bibitem{wordnet}
{\sc Fellbaum, C.}
\newblock Wordnet: An electronic lexical database.
\newblock In {\em Cambridge, MA: MIT Press\/} (1998), pp.~3651--3657.

\bibitem{Entity-centric}
{\sc Godbole, A., Kavarthapu, D., Das, R., Gong, Z., Singhal, A., Zamani, H.,
  Yu, M., Gao, T., Guo, X., Zaheer, M., et~al.}
\newblock Multi-step entity-centric information retrieval for multi-hop
  question answering.
\newblock {\em arXiv preprint arXiv:1909.07598\/} (2019).

\bibitem{he2017dureader}
{\sc He, W., Liu, K., Liu, J., Lyu, Y., Zhao, S., Xiao, X., Liu, Y., Wang, Y.,
  Wu, H., She, Q., et~al.}
\newblock Dureader: a chinese machine reading comprehension dataset from
  real-world applications.
\newblock {\em arXiv preprint arXiv:1711.05073\/} (2017).

\bibitem{ranking-function}
{\sc Htut, P.~M., Bowman, S.~R., and Cho, K.}
\newblock Training a ranking function for open-domain question answering.
\newblock {\em NAACL\/} (2018).

\bibitem{re3}
{\sc Hu, M., Peng, Y., Huang, Z., and Li, D.}
\newblock Retrieve, read, rerank: Towards end-to-end multi-document reading
  comprehension.
\newblock {\em arXiv preprint arXiv:1906.04618\/} (2019).

\bibitem{joshi2017triviaqa}
{\sc Joshi, M., Choi, E., Weld, D.~S., and Zettlemoyer, L.}
\newblock Triviaqa: A large scale distantly supervised challenge dataset for
  reading comprehension.
\newblock {\em arXiv preprint arXiv:1705.03551\/} (2017).

\bibitem{lin2019kagnet}
{\sc Lin, B.~Y., Chen, X., Chen, J., and Ren, X.}
\newblock Kagnet: Knowledge-aware graph networks for commonsense reasoning.
\newblock {\em arXiv preprint arXiv:1909.02151\/} (2019).

\bibitem{dsqa}
{\sc Lin, Y., Ji, H., Liu, Z., and Sun, M.}
\newblock Denoising distantly supervised open-domain question answering.
\newblock In {\em Proceedings of the 56th Annual Meeting of the Association for
  Computational Linguistics (Volume 1: Long Papers)\/} (2018), pp.~1736--1745.

\bibitem{liu2004conceptnet}
{\sc Liu, H., and Singh, P.}
\newblock Conceptnet—a practical commonsense reasoning tool-kit.
\newblock {\em BT technology journal 22}, 4 (2004), 211--226.

\bibitem{minimal}
{\sc Min, S., Zhong, V., Socher, R., and Xiong, C.}
\newblock Efficient and robust question answering from minimal context over
  documents.
\newblock {\em arXiv preprint arXiv:1805.08092\/} (2018).

\bibitem{msmarco}
{\sc Nguyen, T., Rosenberg, M., Song, X., Gao, J., Tiwary, S., Majumder, R.,
  and Deng, L.}
\newblock Ms marco: A human-generated machine reading comprehension dataset.

\bibitem{termquery}
{\sc Ni, J., Zhu, C., Chen, W., and McAuley, J.}
\newblock Learning to attend on essential terms: An enhanced retriever-reader
  model for open-domain question answering.
\newblock In {\em Proceedings of the 2019 Conference of the North American
  Chapter of the Association for Computational Linguistics: Human Language
  Technologies, Volume 1 (Long and Short Papers)\/} (2019), pp.~335--344.

\bibitem{retrieveandread}
{\sc Nishida, K., Saito, I., Otsuka, A., Asano, H., and Tomita, J.}
\newblock Retrieve-and-read: Multi-task learning of information retrieval and
  reading comprehension.
\newblock In {\em Proceedings of the 27th ACM International Conference on
  Information and Knowledge Management\/} (2018), ACM, pp.~647--656.

\bibitem{pang2019has}
{\sc Pang, L., Lan, Y., Guo, J., Xu, J., Su, L., and Cheng, X.}
\newblock Has-qa: Hierarchical answer spans model for open-domain question
  answering.
\newblock {\em arXiv preprint arXiv:1901.03866\/} (2019).

\bibitem{iterativequerygeneration}
{\sc Qi, P., Lin, X., Mehr, L., Wang, Z., and Manning, C.~D.}
\newblock Answering complex open-domain questions through iterative query
  generation.
\newblock {\em EMNLP\/} (2019).

\bibitem{rajpurkar2016squad:}
{\sc Rajpurkar, P., Zhang, J., Lopyrev, K., and Liang, P.}
\newblock Squad: 100,000+ questions for machine comprehension of text.
\newblock {\em EMNLP\/} (2016).

\bibitem{PRP}
{\sc Robertson, S.}
\newblock The probability ranking principle in ir.
\newblock {\em Journal of Documentation 33\/} (12 1977), 294--304.

\bibitem{relationnetwork}
{\sc Santoro, A., Raposo, D., Barrett, D.~G., Malinowski, M., Pascanu, R.,
  Battaglia, P., and Lillicrap, T.}
\newblock A simple neural network module for relational reasoning.
\newblock In {\em Advances in neural information processing systems\/} (2017),
  pp.~4967--4976.

\bibitem{seo2019realtime}
{\sc Seo, M., Lee, J., Kwiatkowski, T., Parikh, A.~P., Farhadi, A., and
  Hajishirzi, H.}
\newblock Real-time open-domain question answering with dense-sparse phrase
  index.
\newblock {\em arXiv preprint arXiv:1906.05807\/} (2019).

\bibitem{fusion}
{\sc Sun, H., Dhingra, B., Zaheer, M., Mazaitis, K., Salakhutdinov, R., and
  Cohen, W.~W.}
\newblock Open domain question answering using early fusion of knowledge bases
  and text.
\newblock {\em arXiv preprint arXiv:1809.00782\/} (2018).

\bibitem{transformer}
{\sc Vaswani, A., Shazeer, N., Parmar, N., Uszkoreit, J., Jones, L., Gomez,
  A.~N., Kaiser, Å., and Polosukhin, I.}
\newblock Attention is all you need.
\newblock {\em Conference on Neural Information Processing Systems\/} (2017),
  6000--6010.

\bibitem{r3}
{\sc Wang, S., Yu, M., Guo, X., Wang, Z., Klinger, T., Zhang, W., Chang, S.,
  Tesauro, G., Zhou, B., and Jiang, J.}
\newblock R 3: Reinforced ranker-reader for open-domain question answering.
\newblock In {\em Thirty-Second AAAI Conference on Artificial Intelligence\/}
  (2018).

\bibitem{reranker}
{\sc Wang, S., Yu, M., Jiang, J., Zhang, W., Guo, X., Chang, S., Wang, Z.,
  Klinger, T., Tesauro, G., and Campbell, M.}
\newblock Evidence aggregation for answer re-ranking in open-domain question
  answering.
\newblock {\em arXiv preprint arXiv:1711.05116\/} (2017).

\bibitem{vnet}
{\sc Wang, Y., Liu, K., Liu, J., He, W., Lyu, Y., Wu, H., Li, S., and Wang, H.}
\newblock Multi-passage machine reading comprehension with cross-passage answer
  verification.
\newblock {\em arXiv preprint arXiv:1805.02220\/} (2018).

\bibitem{wang2018joint}
{\sc Wang, Z., Liu, J., Xiao, X., Lyu, Y., and Wu, T.}
\newblock Joint training of candidate extraction and answer selection for
  reading comprehension.
\newblock {\em arXiv preprint arXiv:1805.06145\/} (2018).

\bibitem{wang2019multibert}
{\sc Wang, Z., Ng, P., Ma, X., Nallapati, R., and Xiang, B.}
\newblock Multi-passage bert: A globally normalized bert model for open-domain
  question answering.
\newblock {\em arXiv preprint arXiv:1908.08167\/} (2019).

\bibitem{bertserini}
{\sc Yang, W., Xie, Y., Lin, A., Li, X., Tan, L., Xiong, K., Li, M., and Lin,
  J.}
\newblock End-to-end open-domain question answering with bertserini.
\newblock {\em arXiv: Computation and Language\/} (2019).

\bibitem{yang2018hotpotqa}
{\sc Yang, Z., Qi, P., Zhang, S., Bengio, Y., Cohen, W.~W., Salakhutdinov, R.,
  and Manning, C.~D.}
\newblock Hotpotqa: A dataset for diverse, explainable multi-hop question
  answering.
\newblock {\em arXiv preprint arXiv:1809.09600\/} (2018).

\end{thebibliography}

\appendix

\end{document}